\def\year{2021}\relax
\documentclass[letterpaper]{article} % DO NOT CHANGE THIS
\usepackage{aaai21}  % DO NOT CHANGE THIS
\usepackage{times}  % DO NOT CHANGE THIS
\usepackage{helvet} % DO NOT CHANGE THIS
\usepackage{courier}  % DO NOT CHANGE THIS
\usepackage[hyphens]{url}  % DO NOT CHANGE THIS
\usepackage{graphicx} % DO NOT CHANGE THIS
\usepackage{natbib}  % DO NOT CHANGE THIS AND DO NOT ADD ANY OPTIONS TO IT
\usepackage{caption} % DO NOT CHANGE THIS AND DO NOT ADD ANY OPTIONS TO IT
\usepackage{arydshln}
\usepackage{multirow}
\usepackage{amsmath}
\usepackage{pdflscape}
\usepackage{longtable}
\usepackage{subcaption}

\title{Beyond Domain APIs: Task-oriented Conversational Modeling with Unstructured Knowledge Access Track in DSTC9}
\author{
  Seokhwan Kim\textsuperscript{\rm 1}, Mihail Eric\textsuperscript{\rm 1}, Behnam Hedayatnia\textsuperscript{\rm 1}, Karthik Gopalakrishnan\textsuperscript{\rm 1},\\
  Yang Liu\textsuperscript{\rm 1}, Chao-Wei Huang\textsuperscript{\rm 2}, Dilek Hakkani-Tur\textsuperscript{\rm 1}\\
}
\affiliations{
  \textsuperscript{\rm 1}Amazon Alexa AI, Sunnyvale, CA, USA\\
  \{seokhwk,mihaeric,behnam,karthgop,yangliud,hakkanit\}@amazon.com\\
  \textsuperscript{\rm 2}National Taiwan University, Taipei, Taiwan\\
  f07922069@csie.ntu.edu.tw
}

\begin{document}

\maketitle

\begin{abstract}
  Most prior work on task-oriented dialogue systems are restricted to a limited coverage of domain APIs, while users oftentimes have domain related requests that are not covered by the APIs.
  This challenge track aims to expand the coverage of task-oriented dialogue systems by incorporating external unstructured knowledge sources.
  We define three tasks: knowledge-seeking turn detection, knowledge selection, and knowledge-grounded response generation.
  We introduce the data sets and the neural baseline models for three tasks.
  The challenge track received a total of 105 entries from 24 participating teams.
  In the evaluation results, the ensemble methods with different large-scale pretrained language models achieved high performances with improved knowledge selection capability and better generalization into unseen data.
\end{abstract}

\section{Introduction}
\label{sec:intro}
Traditionally, task-oriented dialogue systems have focused on providing information and performing actions that can be handled only by given databases or APIs.
However, in addition to task-focused requests, users also have needs that go beyond what is provided by the backend resources.
For example, while most virtual assistants can help users book a hotel, a restaurant or movie tickets, they fall short of answering potential follow-up questions users may have,
such as: where to park vehicles; whether they are allowed to bring pets or children to the reserved place; or what the cancellation policy is.
No API/DB entry is usually available to handle such requests.
On the other hand, relevant domain knowledge is already available on web pages in the form of descriptions, FAQs and customer reviews for many of these out-of-coverage scenarios.
Since current dialogue systems don't incorporate these external knowledge sources into task-oriented conversational modeling,
users need to visit the websites by themselves to find out any additional information beyond API/DB coverage, making conversational interactions inefficient.

Recently, we proposed a new conversational modeling tasks~\cite{kim-etal-2020-beyond} towards frictionless task-oriented scenarios,
where the flow of the conversation does not break when users have requests that are out of the coverage of APIs/DB but potentially are already available in external knowledge sources.
Inspired by recent studies on knowledge-grounded social conversations~\cite{zhou2018commonsense,dinan2018wizard,galley2019grounded,gopalakrishnan2019topical},
we defined the three main tasks: knowledge-seeking turn detection, knowledge selection and knowledge-grounded response generation.
This challenge track focuses on those tasks to develop end-to-end dialogue systems which understand relevant domain knowledge and generate system responses with the selected knowledge.

\section{Task Formulations}
\label{sec:problem}

\begin{figure*}
  \centering
  \includegraphics[width=\textwidth]{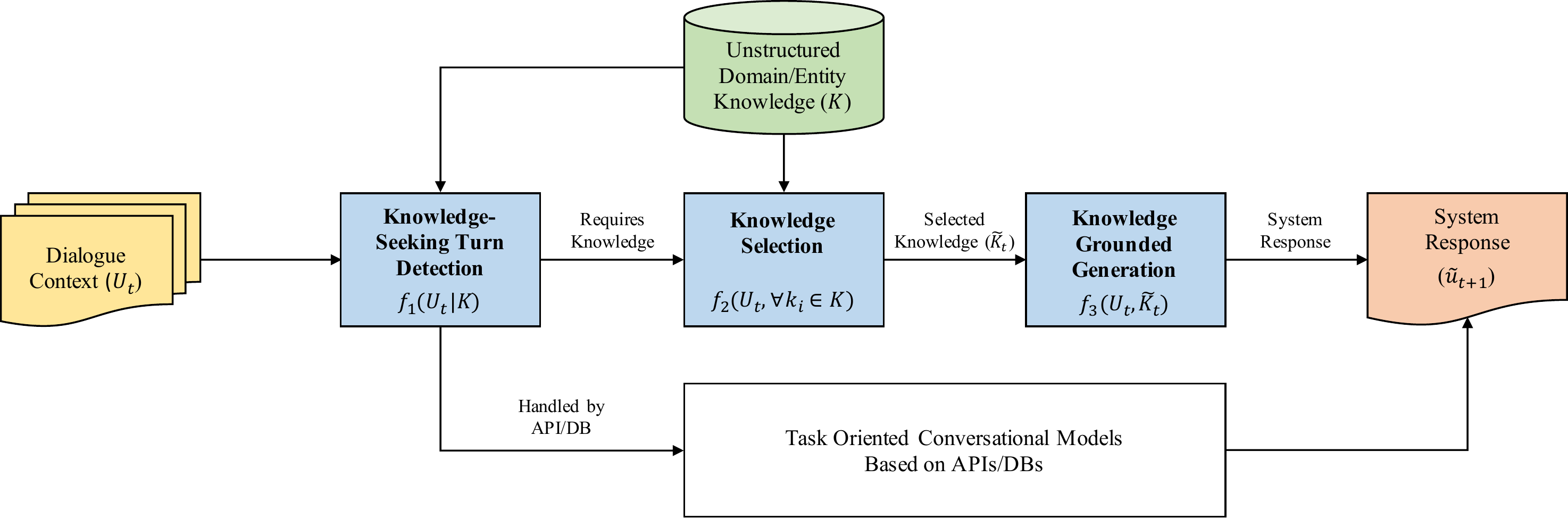}
  \caption{A baseline architecture for task-oriented conversational modeling grounded on unstructured knowledge}
  \label{fig:overview}
\end{figure*}

We define the tasks based on a simple baseline architecture (Figure~\ref{fig:overview})
which decouples turns that could be handled by existing task-oriented conversational models with no extra knowledge
and turns that require external knowledge resources.
We assume that a conventional API-based system already exists and focus on the new knowledge access branch which takes a dialogue context $U_t = \{u_{t-w+1}, \cdots, u_{t-1}, u_t\}$ and knowledge snippets $K = \{k_1, \cdots, k_n\}$,
where $u_{i}$ is the $i$-th utterance in a given dialogue, $t$ is the time-step of the current user utterance to be processed, and $w$ is the dialogue context window size.
In this challenge, we aim to develop systems to generate a context-appropriate system response $\tilde{u}_{t+1}$ grounded on a set of relevant knowledge snippets $\tilde{K} \subset K$.
The remainder of this section presents the detailed formulations of the three main tasks: \textit{`Knowledge-seeking Turn Detection'}, \textit{`Knowledge Selection'}, and \textit{`Knowledge-grounded Response Generation'}.

\subsection{Knowledge-seeking Turn Detection}
\label{sec:problem:task1}

For each given turn at $t$, a system first needs to decide whether to continue an existing API-based scenario or trigger the knowledge access branch.
We call this task \textit{Knowledge-seeking Turn Detection}.
This problem is defined as a binary classification task formulated as follows:
\begin{equation*}
  f_1(U_t|K) = \left\{ \begin{array}{ll}
                       1 & \mbox{if $\exists k \in K$ satisfies $u_t$,}\\
                       0 & \mbox{otherwise,}
                     \end{array} \right.
\end{equation*}
which we assume that every turn can be handled by either branch in the architecture (Figure~\ref{fig:overview}).

\subsection{Knowledge Selection}
\label{sec:problem:task2}

Once a given user turn at $t$ is determined as a knowledge-seeking turn by $f_1(U_t|K)$,
it moves forward with \textit{Knowledge Selection} to sort out the relevant knowledge snippets.
This task takes each pair of $U_t$ and $k_i \in K$ and predicts whether they are relevant or not as follows:
\begin{equation*}
  f_2(U_t, k_i) = \left\{ \begin{array}{ll}
                         1 & \mbox{if $k_i \in K$ is relevant to $U_t$,}\\
                         0 & \mbox{otherwise.}
                       \end{array} \right.
\end{equation*}

Since more than one knowledge snippet can be relevant to a single turn,
we form a task output $\tilde{K}$ including all the positive knowledge snippets from $f_2(U_t, k)$, as follows:
\begin{equation*}
  \tilde{K_t} = \left\{k_i | k_i \in K \wedge f_2(U_t, k_i) = 1 \right\} \subset K.
\end{equation*}

\subsection{Knowledge-grounded Response Generation}
\label{sec:problem:task3}
Finally, a system response $\tilde{u}_{t+1}$ is generated based on both dialogue context $U_t$ and the selected knowledge snippets $\tilde{K_t}$, as follows:
\begin{equation*}
  f_3(U_t, \tilde{K_t}) = \tilde{u}_{t+1}.
\end{equation*}
Each generated response is supposed to provide the user with the requested information grounded on the properly selected knowledge sources.
In addition, the response should be naturally connected to the previous turns.

\section{Data}
\label{sec:data}
This challenge track uses two different data sets (Table~\ref{tbl:data_stats}).
The first data is an augmented version of MultiWOZ 2.1~\cite{eric2019multiwoz} that includes newly introduced knowledge-seeking turns in the MultiWOZ conversations.
The data augmentation was incrementally done by the crowdsourcing tasks described in~\cite{kim-etal-2020-beyond}.
A total of 22,834 utterance pairs were newly collected based on 2,900 knowledge candidates from the FAQ webpages about the domains and the entities in MultiWOZ databases.
For the challenge track, we divided the whole data into three subsets: train, validation and test.
The first two sets were released in the development phase along with the ground-truth annotations and human responses for participants to develop their models.

In the evaluation phase, we released the test split of the augmented MultiWOZ 2.1 and the other conversations collected from scratch about touristic information for San Francisco.
To evaluate the generalizability of models, the new conversations cover knowledge, locale and domains that are unseen from the train and validation data sets.
In addition, this test set includes not only written conversations, but also spoken dialogs to evaluate system performance across different modalities~\cite{gopalakrishnan2020neural}.
All the backend resources for this data collection were also released, which includes 9,139 knowledge snippets and 855 database entries for San Francisco.

\begin{table}[t]
  \centering
  \caption{Statistics of the challenge track data sets}
  \label{tbl:data_stats}
  \small
  \begin{tabular}{l l r r r}
  \hline
    & & \# & total \# & \# knowledge \\
    Source & Split & dialogs & instances & seeking turns \\ \hline
    \multirow{3}{*}{MultiWOZ} & Train & 7,190 & 71,348 & 19,184 \\
    & Valid & 1,000 & 9,663 & 2,673 \\
    & Test & 977 & 2,084 & 977 \\ \hdashline[.4pt/1pt]
    \multirow{2}{*}{SF} & Written & 900 & 1,834 & 900\\
    & Spoken & 107 & 263 & 104\\ \hline
    % Total & 10,438 & 9,072 & 161,192
  \end{tabular}
\end{table}

\section{Baseline}
\label{sec:baseline}

\begin{figure*}
  \centering
  \begin{subfigure}[b]{0.20\textwidth}
    \includegraphics[height=2in]{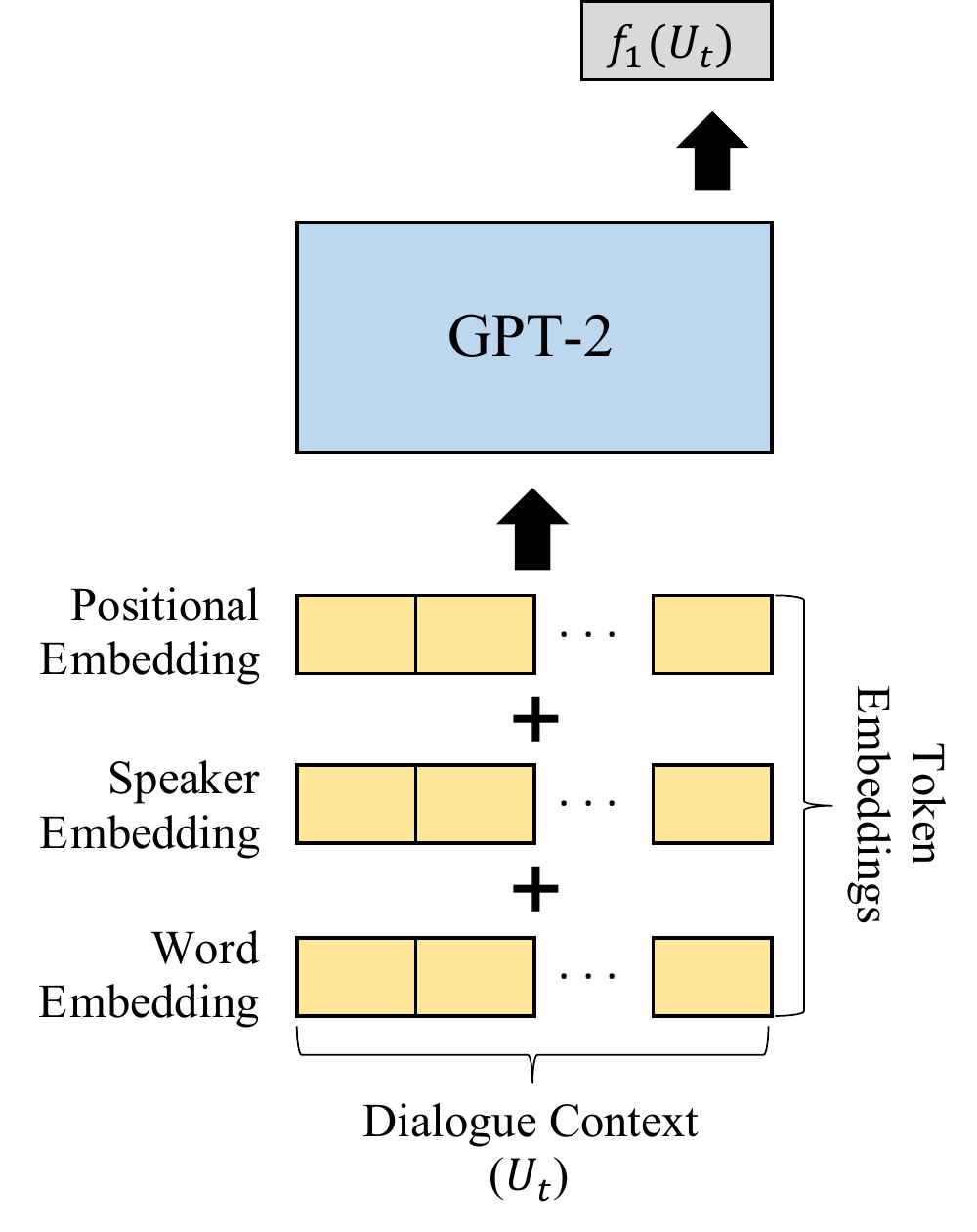}
    \caption{Detection baseline}
    \label{fig:baseline_task1}
  \end{subfigure}
  ~
  \begin{subfigure}[b]{0.31\textwidth}
    \includegraphics[height=2in]{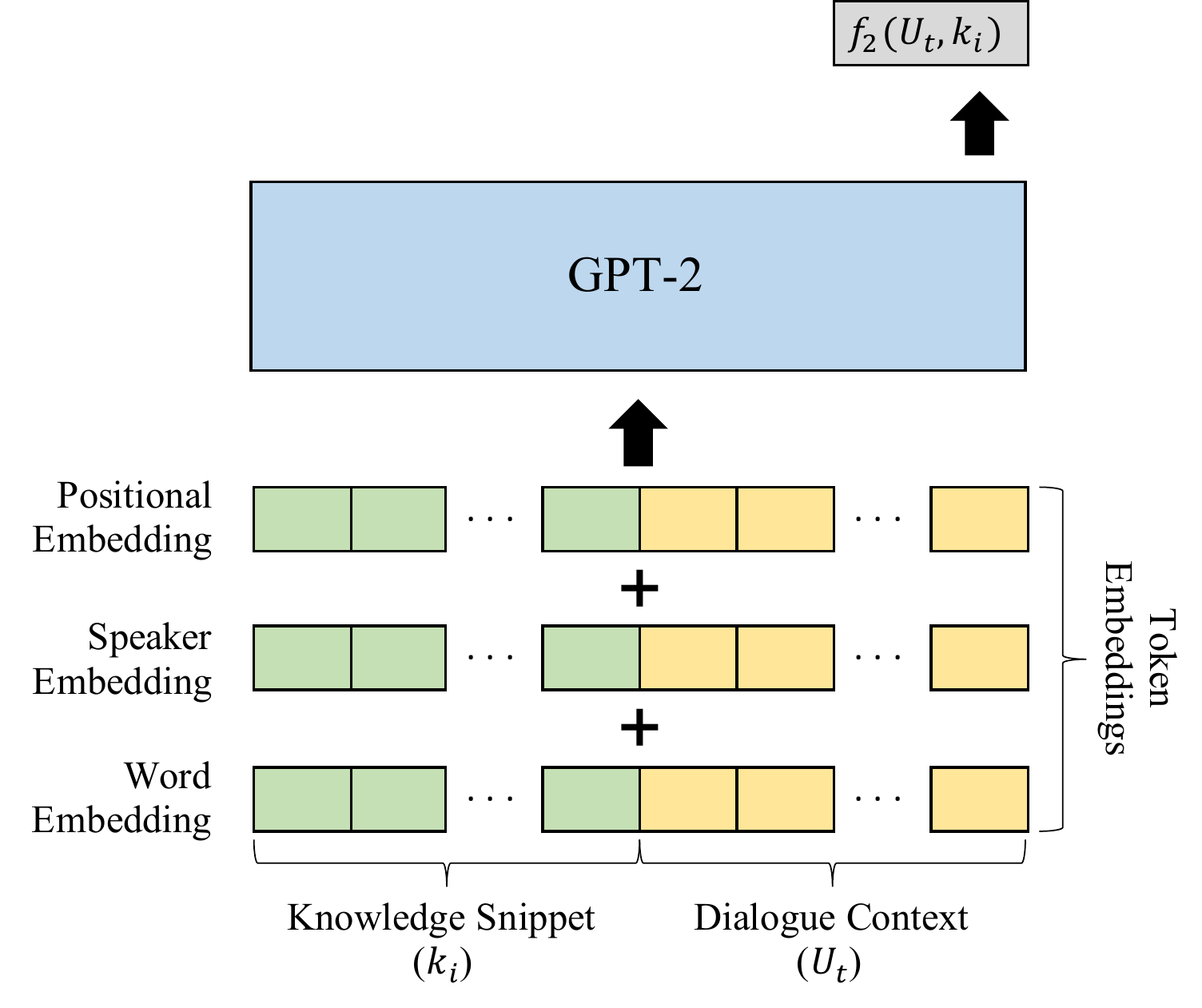}
    \caption{Selection baseline}
    \label{fig:baseline_task2}
  \end{subfigure}
  ~
  \begin{subfigure}[b]{0.44\textwidth}
    \includegraphics[height=2in]{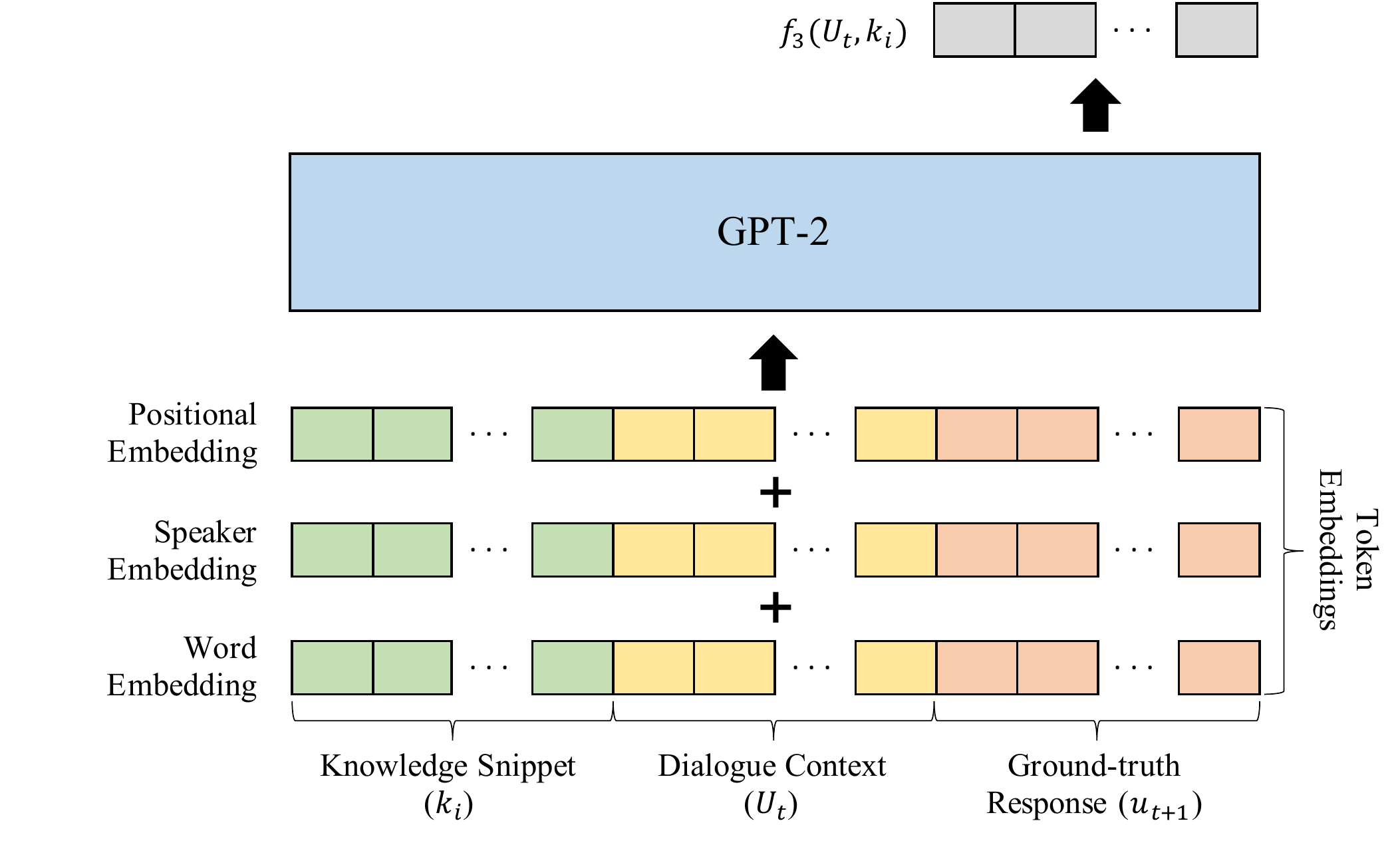}
    \caption{Generation baseline}
    \label{fig:baseline_task3}
  \end{subfigure}
  \caption{Baseline models for three tasks}\label{fig:baselines}
\end{figure*}  

In addition to the data, we released the neural baseline models for all the challenge track tasks.
We fine-tuned the pre-trained GPT-2~\cite{radford2019language} \textit{small} model separately for each task.
Specifically, we first trained a binary classifier (Figure~\ref{fig:baseline_task1}) for the knowledge-seeking turn detection task.
The model takes the dialogue context $U_t$ as input and generates the dialogue representation as the final layer output for $[EOS]$ which is a special token appended to the end of the input sequence.
We added a single layer feed-forward network on top of the dialogue representation and trained it with binary cross-entropy loss.

For the knowledge selection task, we trained another binary classification model (Figure~\ref{fig:baseline_task2}) over a pair of encoded texts as is done in prior Transformer sentence relationship models~\cite{Nogueira2019PassageRW}.
The model takes the concatenation of the utterances in $U_t$ and the sentences in $k_i$ as an input instance.
We use the final layer output at the same position to the $[EOS]$ token as input to a single layer feed-forward network to obtain a probability $s_i$ that $k_i$ is relevant to the given dialogue context $U_t$.
The model fine-tuning was done also with the binary cross-entropy loss as follows:
\begin{equation*}
  L = -\displaystyle\sum_{i\in I_{pos}} \log(s_i) - \displaystyle\sum_{i\in I_{neg}} \log(1 - s_i),
\end{equation*}
where $I_{pos}$ refers to the set of knowledges that are relevant to the given dialogue context and $I_{neg}$ refers to those that are not.
The negative candidates are sampled uniformly at random from the entire set of knowledge snippets.

Finally, we developed a neural response generation baseline also by fine-tuning the GPT-2 \textit{small} model with a standard language modeling objective on our data set.
The model takes ground-truth knowledge snippets concatenated to each input dialog context (Figure~\ref{fig:baseline_task3}) for fine-tuning.

All the three models were implemented using the \textit{transformers} library~\cite{Wolf2019HuggingFacesTS}~\footnote{https://huggingface.co/transformers/}.
We fine-tuned the baseline models for a fixed number of 10 epochs with a truncation window of 128 tokens for both dialog context $U_t$ and knowledge snippet $k_i$, as default.

\section{Evaluation Criteria}

\begin{table}[t]
  \centering
  \caption{Objective evaluation metrics}
  \small
  \begin{tabular}{l l}
  \hline
    Task & Metrics \\ \hline
    Task \#1 & Precision/Recall/F-measure \\\hdashline[.4pt/1pt]
    Task \#2 & MRR@5, Recall@1, Recall@5 \\ \hdashline[.4pt/1pt]
    Task \#3 & BLEU-1, BLEU-2, BLEU-3, BLEU-4, METEOR\\
    & ROUGE-1, ROUGE-2, ROUGE-L \\ \hline
  \end{tabular}
  \label{tbl:track1_metrics}
\end{table}

Each participating team submitted up to five system outputs each of which contains the results for all three tasks on the unlabeled test instances.
We first evaluated each submission using the task-specific objective metrics (Table~\ref{tbl:track1_metrics}) by comparing to the ground-truth labels and responses.
Considering the dependencies between the tasks in the pipelined architecture, the final scores for knowledge selection and knowledge-grounded response generation are computed by considering the first step  knowledge-seeking turn detection recall and precision performance, as follows:
\begin{equation*}
  \tilde{f_1}(x) = \left\{ \begin{array}{ll}
    1 & \mbox{if $x$ is predicted as a knowledge-seeking turn,}\\
    0 & \mbox{otherwise}
  \end{array} \right.
\end{equation*}

\begin{equation*}
  S_p(X) = \frac{\sum_{x_i \in X}\left(s(x_i) \cdot f_1(x_i) \cdot \tilde{f_1}(x_i)\right)}{\sum_{x_i \in X}\tilde{f_1}(x_i)},
\end{equation*}
\begin{equation*}
  S_r(X) = \frac{\sum_{x_i \in X}\left(s(x_i) \cdot f_1(x_i) \cdot \tilde{f_1}(x_i)\right)}{\sum_{x_i \in X}f_1(x_i)},
\end{equation*}
\begin{equation}
  S_f(X) = \frac{2 \cdot S_p(X) \cdot S_r(X)}{S_p(X) + S_r(X)},
  \label{eq:end_to_end_score}
\end{equation}
where $s(x)$ is the knowledge selection or response generation score in a target metric for a single instance $x \in X$.

Then, we aggregated a set of multiple scores across different tasks and metrics into a single overall score computed by the mean reciprocal rank, as follows:
\begin{equation}
  S_{overall}(e) = \frac{1}{|M|}\sum_{i=1}^{|M|}\frac{1}{rank_i(e)},
  \label{eq:overall_score}
\end{equation}
where $rank_i(e)$ is the ranking of the submitted entry $e$ in the $i$-th metric against all the other submissions and $M$ is the number of metrics we considered.

\begin{figure*}
  \centering
  \begin{subfigure}[b]{0.48\textwidth}
    \includegraphics[width=\linewidth]{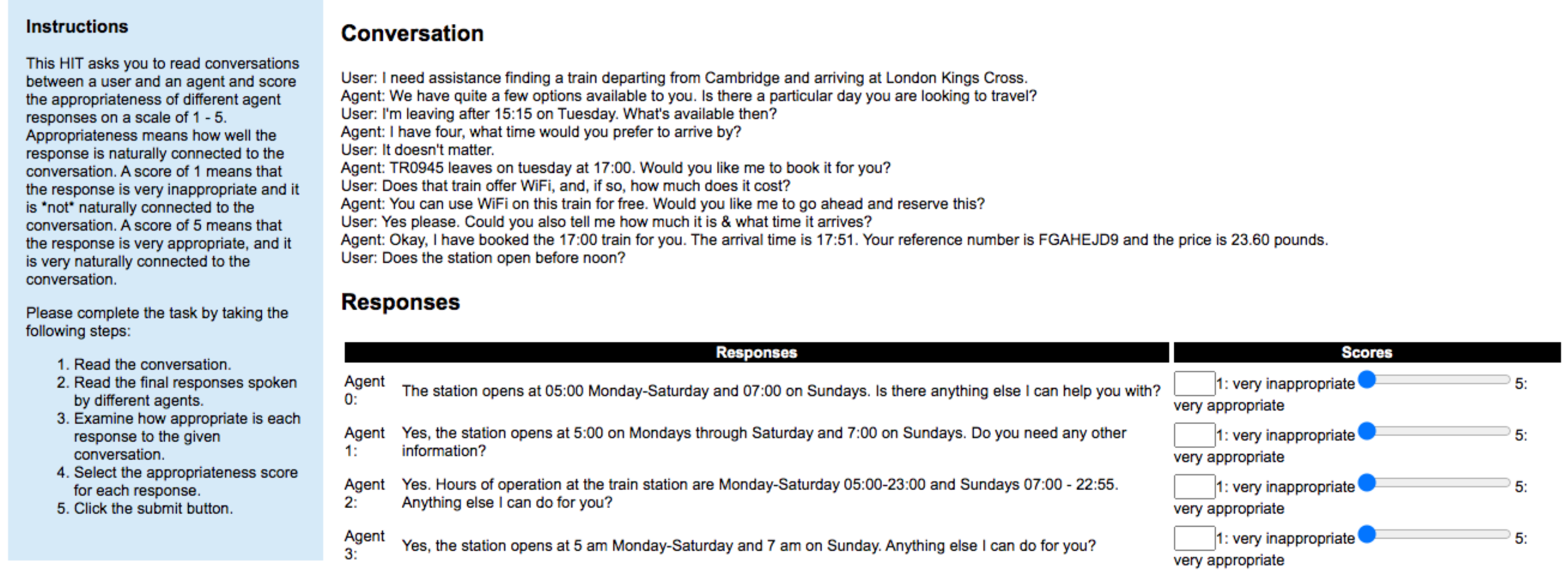}
    \caption{Appropriateness}
    \label{fig:human_eval_appropriateness}
  \end{subfigure}
  \begin{subfigure}[b]{0.48\textwidth}
    \includegraphics[width=\linewidth]{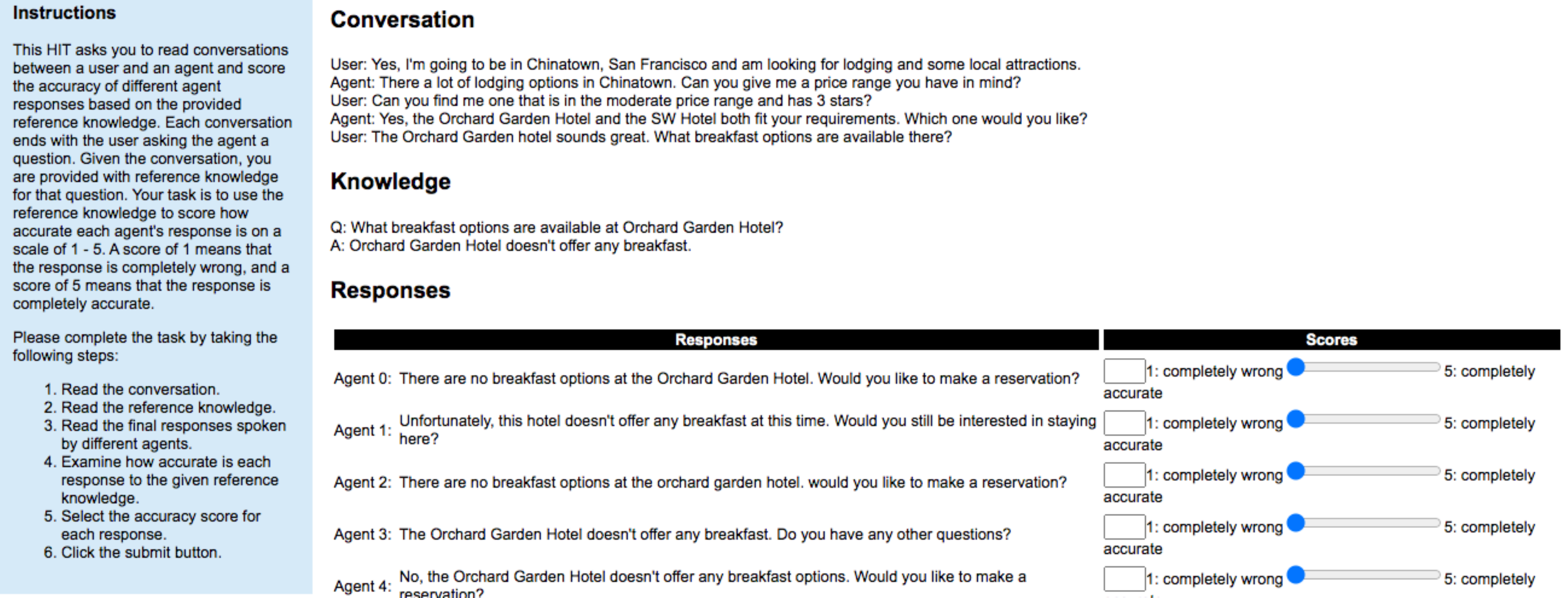}
    \caption{Accuracy}
    \label{fig:human_eval_accuracy}
  \end{subfigure}
  \caption{Crowdsourcing user interface for the human evaluations}\label{fig:human_eval}
\end{figure*}  

Based on the overall objective score, we selected the finalists to be manually evaluated by the following two crowd sourcing tasks:
\begin{itemize}
\item Appropriateness: This task asks crowd workers to score how well a system output is naturally connected to a given conversation on a scale of 1-5.
\item Accuracy: This task asks crowd workers to score the accuracy of a system output based on the provided reference knowledge on a scale of 1-5.
\end{itemize}
Figure~\ref{fig:human_eval} shows the crowdsourcing user interfaces for both human evaluation tasks.
While the Accuracy task includes the ground-truth knowledge snippet for crowd workers to compare it with each response,
the Appropriateness interface provides a conversation and responses only to evaluate the coherency of each response regardless of the factual correctness to the knowledge.
In both tasks, we assigned each instance to three crowd workers and took their average as the final human evaluation score for the instance.
Those scores were then aggregated over the entire test set following Equation~\ref{eq:end_to_end_score}, i.e., weighted by the knowledge-seeking turn detection performance.
Finally, we used the average of the Appropriateness and Accuracy scores to determine the official ranking of the systems in the challenge track.

\section{Results}
We received 105 entries in total submitted from 24 participating teams.
To preserve anonymity, the teams were identified by numbers from 1 to 24.

\subsection{Objective Evaluation Results}
Table~\ref{tbl:track1_objective} shows the objective evaluation results of the best entry from each team selected based on the overall score (Equation~\ref{eq:overall_score}).
The full results including all the submitted entries are available on the track repository\footnote{\url{https://github.com/alexa/alexa-with-dstc9-track1-dataset}}.

\begin{figure}
  \includegraphics[width=\linewidth]{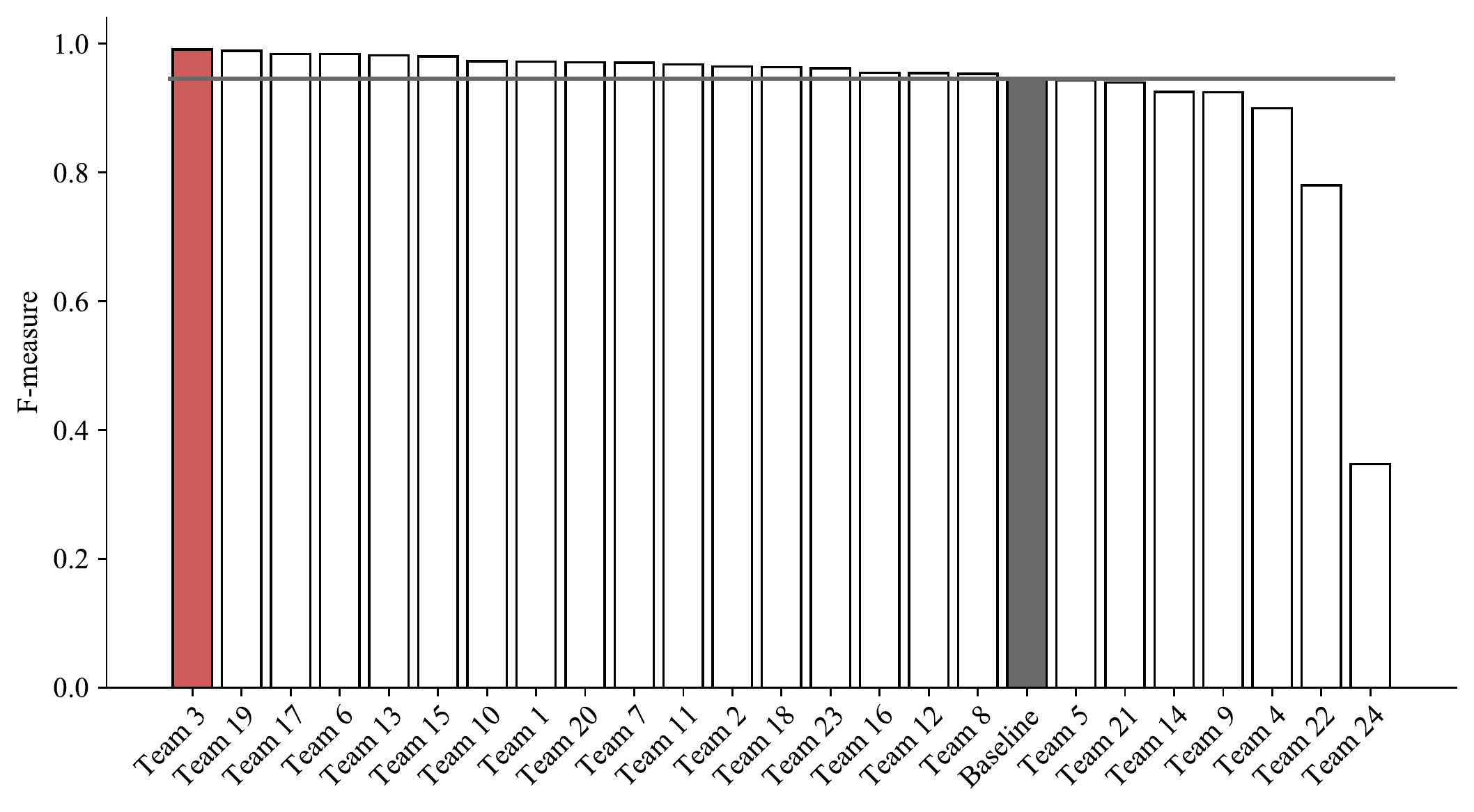}
  \caption{Comparisons of the knowledge-seeking turn detection performances in F-measure. The horizontal line indicates the baseline performance.}
  \label{fig:result_task1}
\end{figure}

Most entries show the improved performances from the baseline models in all three tasks.
As shown in Figure~\ref{fig:result_task1}, 17 teams achieved higher knowledge-seeking turn detection performances than the baseline classifier.
Especially, Team 3 reached over 99\% F-measure in their detection results by ensemble of four difference models including UniLM~\cite{dong2019unified} and three RoBERTa~\cite{liu2019roberta} variants.

\begin{figure}
  \centering
  \includegraphics[width=\linewidth]{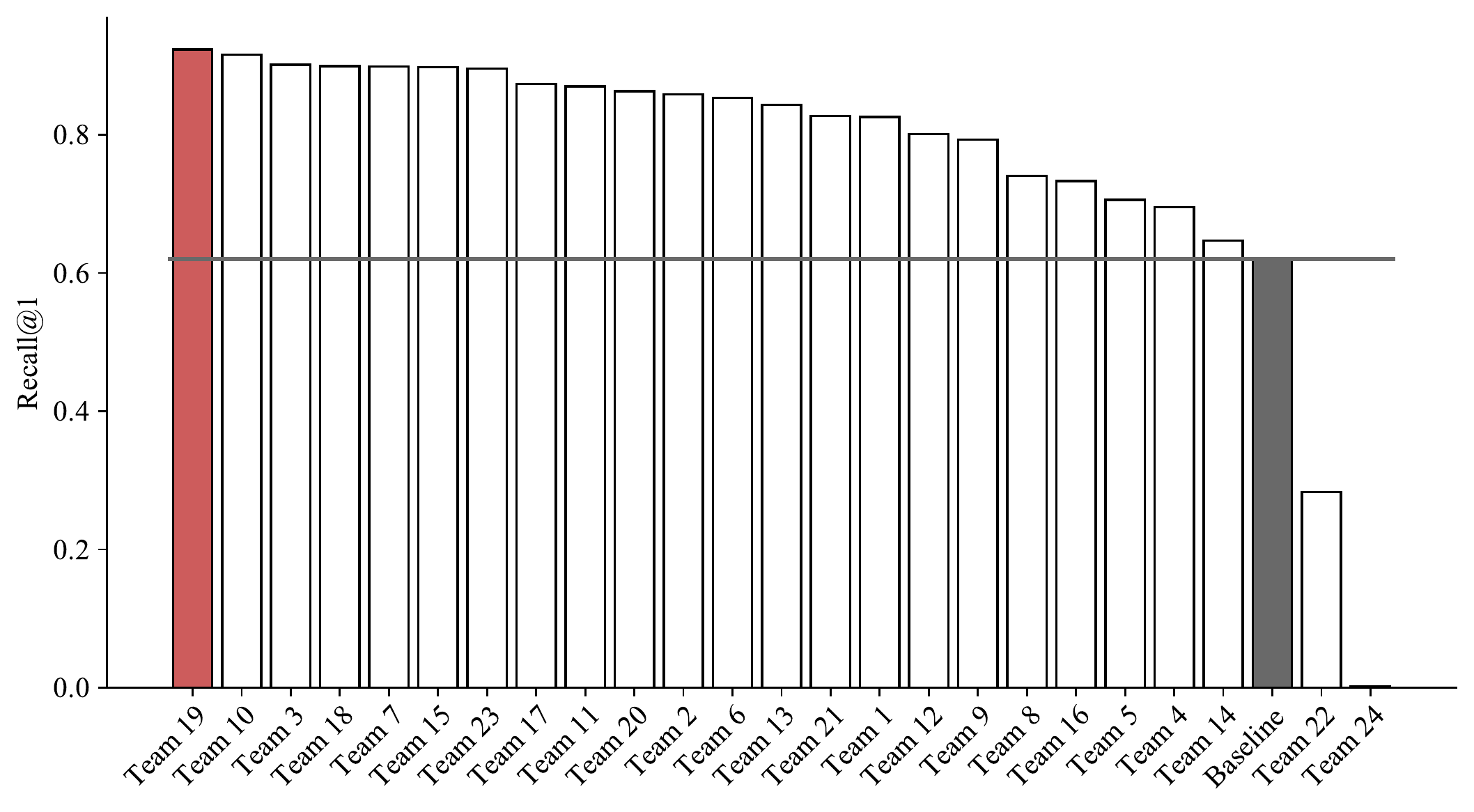}
  \caption{Comparisons of the knowledge selection performances in Recall@1. The horizontal line indicates the baseline performance.}
  \label{fig:result_task2}
\end{figure}

The knowledge selection was the most significantly improved task by the participating teams in this challenge track.
Among 22 teams submitted better knowledge selection results than the baseline (Figure~\ref{fig:result_task2}),
Team 19 was the best in MRR@5 and Recall@1 when they used a model ensemble approach for knowledge selection.
Their best entries achieved 92.35\% in Recall@1 which is over 30\% higher than the baseline and also 0.77\% higher than the second best team's results.

\begin{figure}
  \centering
  \includegraphics[width=\linewidth]{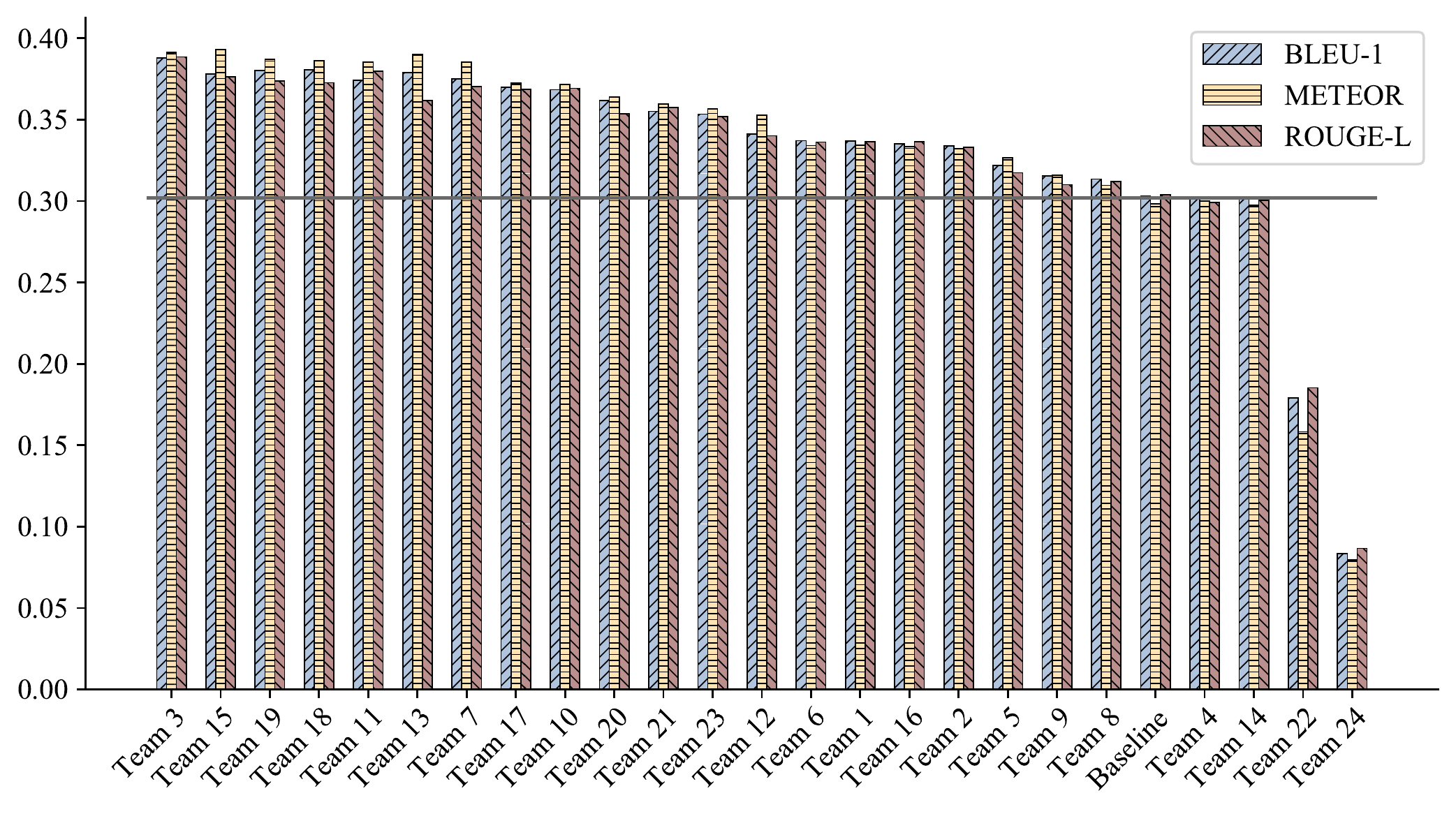}
  \caption{Comparisons of the Knowledge-grounded generation performances in the featured metrics. The horizontal line indicates the average of the baseline scores.}
  \label{fig:result_task3}
\end{figure}

For the knowledge-grounded response generation task (Figure~\ref{fig:result_task3}), Team 3 got the highest scores in six out of eight objective metrics, while Team 15 was better in the other two metrics: BLEU-3 and METEOR.
The best entry from Team 3 combines nine systems including GPT-2~\cite{radford2019language}, DialoGPT\cite{zhang-etal-2020-dialogpt}, and UniLM~\cite{dong2019unified} using Minimum Bayes Risk decoding~\cite{goel2000minimum}.
Team 15 also employed two different pre-trained language models, Electra~\cite{clark2020electra} and RoBERTa~\cite{liu2019roberta} and developed a ranking policy on multiple model outputs. 

\begin{landscape}
\begin{table}
  \caption{Objective evaluation results of the best entries from each team. Bold denotes the best score for each metric; underline indicates the best results among the single models with no ensemble; and $^*$ indicates the finalists selected for the human evaluations.}
  \label{tbl:track1_objective}
  \centering
  \small
  \begin{tabular}{l l l l l l l l l l l l l l l l l l}
    & \multicolumn{3}{l}{Task \#1: Detection} & \multicolumn{3}{l}{Task \#2: Selection} & \multicolumn{8}{l}{Task \#3: Generation}\\
    Team-Entry & P & R & F & MRR@5 & R@1 & R@5 & BLEU-1 & BLEU-2 & BLEU-3 & BLEU-4 & METEOR & ROUGE-1 & ROUGE-2 & ROUGE-L\\ \hline
Baseline & 0.9933 & 0.9021 & 0.9455 & 0.7263 & 0.6201 & 0.8772 & 0.3031 & 0.1732 & 0.1005 & 0.0655 & 0.2983 & 0.3386 & 0.1364 & 0.3039 \\ \hdashline[.4pt/1pt]
1 - 1 & 0.9670 & 0.9773 & 0.9721 & 0.8786 & 0.8255 & 0.9435 & 0.3333 & 0.1994 & 0.1186 & 0.0750 & 0.3296 & 0.3728 & 0.1578 & 0.3305 \\
2 - 2 & 0.9936 & 0.9369 & 0.9644 & 0.8914 & 0.8527 & 0.9509 & 0.3338 & 0.1999 & 0.1179 & 0.0755 & 0.3322 & 0.3751 & 0.1583 & 0.3329 \\
3 - 0 & 0.9964 & 0.9859 & 0.9911 & 0.9395 & 0.9013 & \bf{0.9840} & \bf{0.3879} & 0.2512 & 0.1658 & 0.1158 & 0.3885 & \bf{0.4335} & 0.2061 & 0.3867 \\
3 - 1$^*$ & 0.9964 & \bf{0.9859} & \bf{0.9911} & 0.9395 & 0.9013 & \bf{0.9840} & 0.3864 & \bf{0.2539} & 0.1692 & \bf{0.1190} & 0.3914 & 0.4332 & \bf{0.2115} & \bf{0.3885} \\
3 - 4 & 0.9964 & 0.9859 & 0.9911 & 0.9395 & 0.9013 & \bf{0.9840} & 0.3766 & 0.2449 & 0.1601 & 0.1113 & 0.3795 & \underline{0.4229} & 0.2036 & \underline{0.3801} \\
4 - 1 & \bf{0.9994} & 0.8183 & 0.8998 & 0.7189 & 0.6950 & 0.7705 & 0.3025 & 0.1782 & 0.1058 & 0.0671 & 0.3001 & 0.3366 & 0.1395 & 0.2990 \\
5 - 2 & 0.9916 & 0.8985 & 0.9428 & 0.7588 & 0.7055 & 0.8337 & 0.3210 & 0.1936 & 0.1175 & 0.0742 & 0.3260 & 0.3572 & 0.1528 & 0.3179 \\
6 - 2 & 0.9838 & 0.9838 & 0.9838 & 0.8842 & 0.8486 & 0.9465 & 0.3362 & 0.1995 & 0.1186 & 0.0763 & 0.3330 & 0.3761 & 0.1580 & 0.3334 \\
7 - 4$^*$ & 0.9957 & 0.9460 & 0.9702 & \underline{0.9309} & 0.8988 & \underline{0.9666} & 0.3752 & 0.2426 & 0.1568 & 0.1050 & 0.3852 & 0.4154 & 0.1957 & 0.3702 \\
8 - 4 & 0.9875 & 0.9207 & 0.9530 & 0.7876 & 0.7403 & 0.8563 & 0.3135 & 0.1839 & 0.1082 & 0.0699 & 0.3097 & 0.3501 & 0.1454 & 0.3119 \\
9 - 1 & 0.9925 & 0.8647 & 0.9242 & 0.8128 & 0.7882 & 0.8508 & 0.3154 & 0.1908 & 0.1123 & 0.0713 & 0.3159 & 0.3517 & 0.1507 & 0.3101 \\
10 - 0$^*$ & 0.9860 & 0.9596 & 0.9726 & 0.9400 & 0.9158 & 0.9670 & 0.3684 & 0.2374 & 0.1531 & 0.1030 & 0.3719 & 0.4113 & 0.1938 & 0.3692 \\
11 - 3$^*$ & 0.9879 & 0.9480 & 0.9675 & 0.9005 & 0.8702 & 0.9377 & 0.3743 & 0.2491 & 0.1693 & 0.1157 & 0.3854 & 0.4179 & 0.2080 & 0.3797 \\
12 - 4 & 0.9951 & 0.9162 & 0.9540 & 0.8395 & 0.8011 & 0.8925 & 0.3374 & 0.2131 & 0.1376 & 0.0885 & 0.3526 & 0.3780 & 0.1736 & 0.3376 \\
13 - 3$^*$ & 0.9794 & 0.9844 & 0.9819 & 0.8832 & 0.8434 & 0.9426 & 0.3787 & 0.2396 & 0.1448 & 0.0985 & 0.3902 & 0.4211 & 0.1894 & 0.3619 \\
14 - 0 & 0.9988 & 0.8617 & 0.9252 & 0.7404 & 0.6466 & 0.8667 & 0.3019 & 0.1738 & 0.1009 & 0.0639 & 0.2974 & 0.3367 & 0.1348 & 0.3003 \\
15 - 3$^*$ & 0.9933 & 0.9677 & 0.9803 & 0.9195 & 0.8975 & 0.9460 & 0.3779 & \underline{0.2532} & \underline{\bf{0.1731}} & \underline{0.1175} & \underline{\bf{0.3931}} & 0.4204 & \underline{0.2113} & 0.3765 \\
16 - 3 & 0.9929 & 0.9197 & 0.9549 & 0.7891 & 0.7327 & 0.8721 & 0.3351 & 0.2066 & 0.1288 & 0.0857 & 0.3334 & 0.3743 & 0.1659 & 0.3364 \\
17 - 0$^*$ & 0.9933 & 0.9748 & \underline{0.9839} & 0.9093 & 0.8713 & 0.9605 & 0.3699 & 0.2394 & 0.1520 & 0.1032 & 0.3724 & 0.4164 & 0.1966 & 0.3687 \\
18 - 3$^*$ & 0.9962 & 0.9329 & 0.9635 & 0.9155 & \underline{0.8994} & 0.9343 & \underline{0.3794} & 0.2455 & 0.1612 & 0.1081 & 0.3864 & 0.4164 & 0.1976 & 0.3707 \\
19 - 2$^*$ & 0.9954 & 0.9818 & 0.9886 & \bf{0.9504} & \bf{0.9235} & 0.9814 & 0.3803 & 0.2449 & 0.1590 & 0.1081 & 0.3869 & 0.4192 & 0.1976 & 0.3738 \\
20 - 4$^*$ & 0.9926 & 0.9505 & 0.9711 & 0.8940 & 0.8628 & 0.9345 & 0.3619 & 0.2269 & 0.1406 & 0.0964 & 0.3637 & 0.3979 & 0.1799 & 0.3535 \\
21 - 3$^*$ & 0.9927 & 0.8920 & 0.9396 & 0.8530 & 0.8269 & 0.8955 & 0.3551 & 0.2300 & 0.1532 & 0.1040 & 0.3594 & 0.3976 & 0.1907 & 0.3570 \\
22 - 2 & 0.9992 & 0.6401 & 0.7803 & 0.3337 & 0.2640 & 0.4671 & 0.1792 & 0.0817 & 0.0435 & 0.0345 & 0.1583 & 0.2033 & 0.0664 & 0.1852 \\
23 - 0$^*$ & 0.9984 & 0.9278 & 0.9618 & 0.9233 & 0.8959 & 0.9555 & 0.3523 & 0.2225 & 0.1431 & 0.0929 & 0.3527 & 0.3927 & 0.1806 & 0.3500 \\
24 - 0 & 0.9882 & 0.2105 & 0.3471 & 0.0038 & 0.0017 & 0.0067 & 0.0835 & 0.0413 & 0.0196 & 0.0123 & 0.0796 & 0.0963 & 0.0322 & 0.0866 \\    
  \end{tabular}
\end{table}
\end{landscape}

For all three sub-tasks, most top systems include multiple models and combine the results by various ensemble methods, which achieved the highest score for all the metrics except BLEU-3 and METEOR for the response generation task, as shown also in Table~\ref{tbl:track1_objective}.

\subsection{Human Evaluation Results}
We selected 12 finalists to be manually evaluated, corresponding to the best entry from each of the top 12 teams in the overall objective score (Equation~\ref{eq:overall_score}).
Table~\ref{tbl:track1_human} shows the official ranking of the finalists based on the human evaluation results.
Team 19 won the challenge track with the highest scores for both Accuracy and Appropriateness, most likely because of their better performance in the knowledge selection task (as in Figure~\ref{fig:result_task2}).

\begin{table}[t]
  \caption{Human evaluation results. Bold indicates the best score for each metric and $\dagger$ denotes the statistical significance ($p<0.01$) from the paired t-test.}
  \label{tbl:track1_human}
  \centering
  \small
  \begin{tabular}{l l l c c c}
  \hline
    Rank & Team & Entry & Accuracy & Appropriateness & Average \\ \hline
    & \multicolumn{2}{l}{Ground-truth} & 4.5930 & 4.4513 & 4.5221  \\ \hdashline[.4pt/1pt]
    1 & 19 & 2 & \bf{4.3917} & \bf{4.3922}$^\dagger$ & \bf{4.3920}$^\dagger$  \\
    2 & 3 & 1 & 4.3480 & 4.3634 & 4.3557  \\
    3 & 10 & 0 & 4.3544 & 4.3201 & 4.3373  \\
    4 & 15 & 3 & 4.3793 & 4.2755 & 4.3274  \\
    5 & 17 & 0 & 4.3360 & 4.3076 & 4.3218  \\
    6 & 7 & 4 & 4.3308 & 4.2989 & 4.3149  \\
    7 & 18 & 3 & 4.3309 & 4.2859 & 4.3084  \\
    8 & 13 & 3 & 4.3763 & 4.2360 & 4.3061  \\
    9 & 23 & 0 & 4.3082 & 4.2665 & 4.2874  \\
    10 & 11 & 3 & 4.2722 & 4.2619 & 4.2670  \\
    11 & 20 & 4 & 4.2283 & 4.2486 & 4.2384  \\
    12 & 21 & 3 & 4.1060 & 4.1560 & 4.1310  \\ \hdashline[.4pt/1pt]
    & \multicolumn{2}{l}{Baseline} & 3.7155 & 3.9386 & 3.8271  \\
    \hline
  \end{tabular}
\end{table}

\begin{figure}[t]
  \centering
  \includegraphics[width=\linewidth]{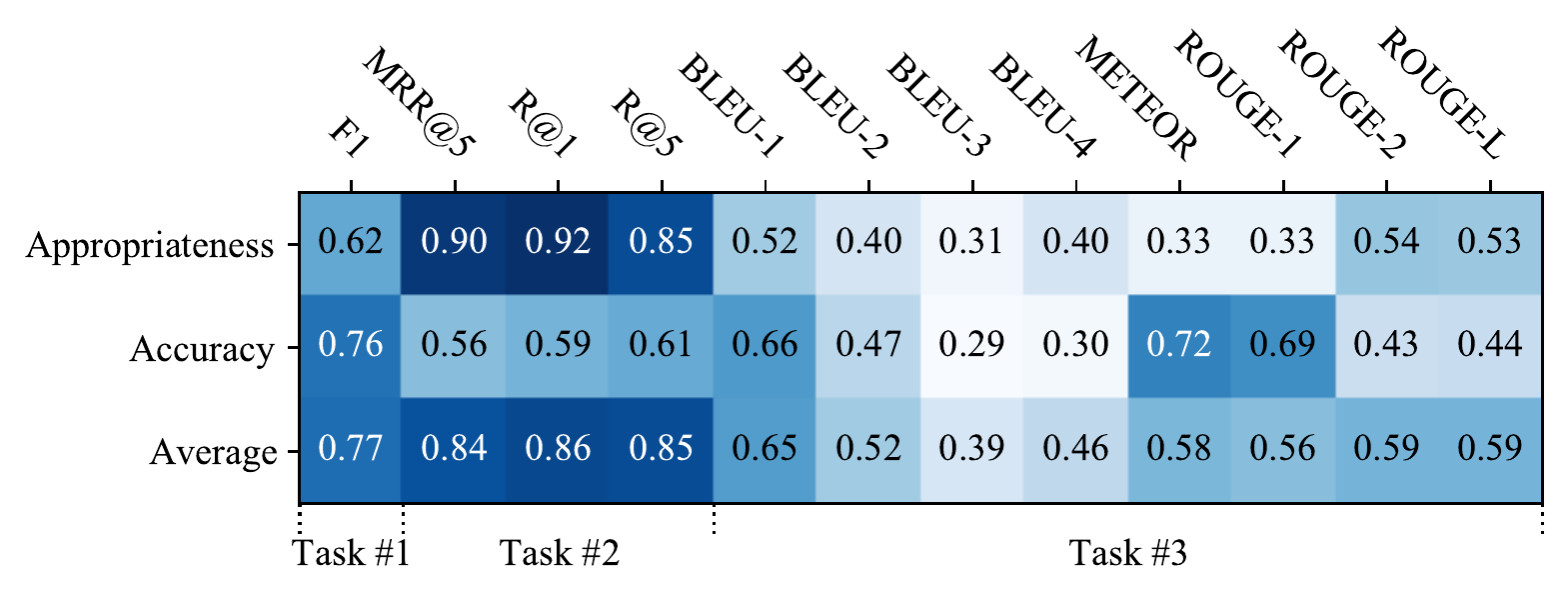}
  \caption{Correlations between the objective and human evaluation metrics in Spearman's $\rho$. The higher score of a pair of metrics has, the stronger correlation they have.}
  \label{fig:metrics_corr}
\end{figure}

The importance of the knowledge selection results towards end-to-end performance is shown also in the correlations between the metrics (Figure~\ref{fig:metrics_corr}).
We calculated the Spearman's rank correlation coefficient~\cite{spearman1961proof} of the ranked lists of all the entries in every pair of objective and human evaluation metrics.
As a results, the knowledge selection metrics showed stronger correlations than the other metrics to the final ranking.
Especially, Recall@1 for the knowledge selection task has the highest correlation to the averaged human evaluation ranking at 0.8601, which is significantly stronger than 0.7692 and 0.6503 with F-measure for the knowledge-seeking turn detection and BLEU-1 for the response generation, respectively.
This implies that the knowledge selection is a key task to improve end-to-end performance.

\begin{figure}[t]
  \centering
  \includegraphics[width=\linewidth]{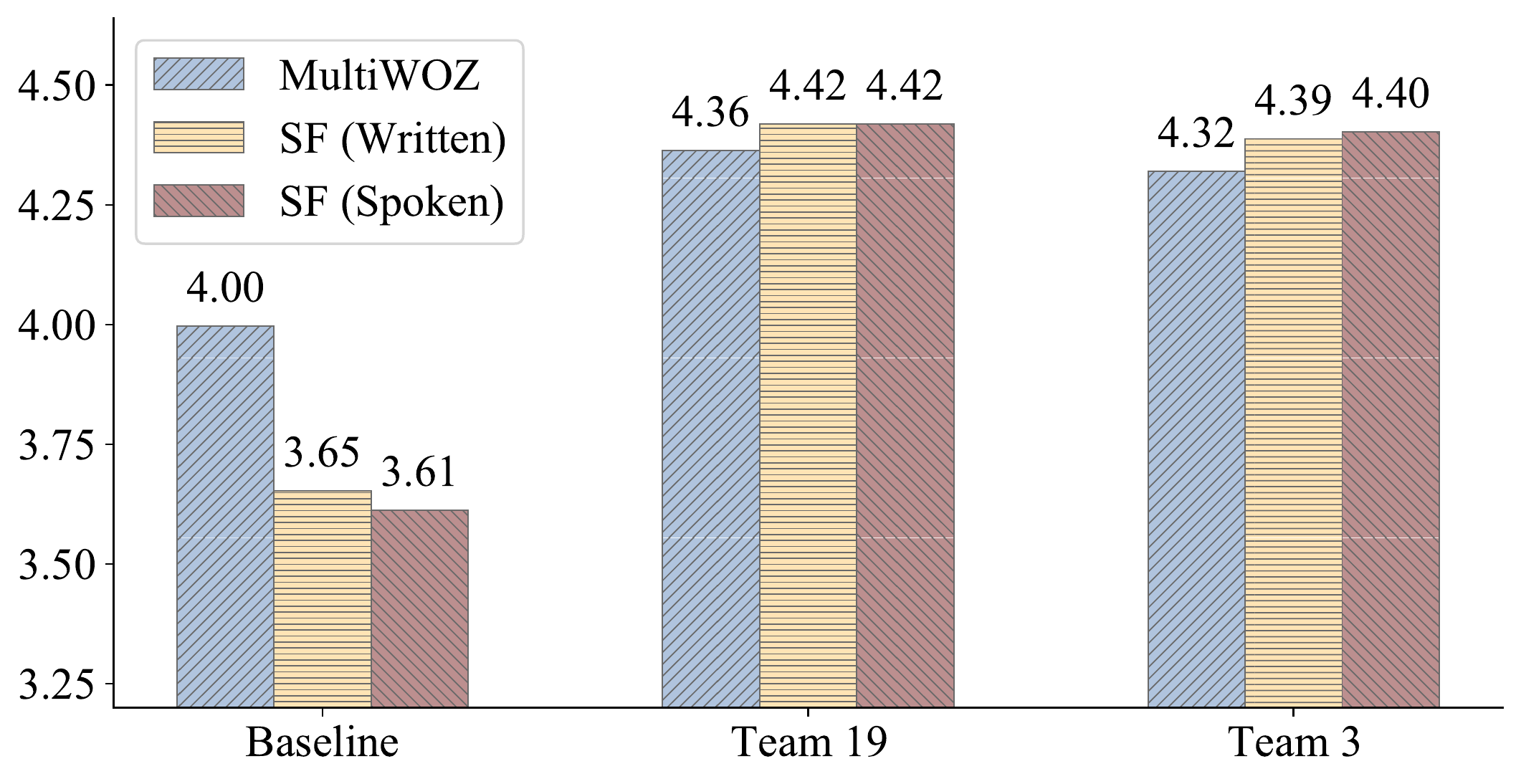}
  \caption{Comparisons of the averaged human evaluation scores between the baseline and the top entries per subset.}
  \label{fig:results_per_subset}
\end{figure}

Figure~\ref{fig:results_per_subset} compares the averaged human evaluation scores between the baseline and the systems from the top-2 teams for each subset of the test data.
While the baseline system shows substantial performance degradation for the unseen subset, the participants' systems have even higher scores for the San Francisco data in both modalities than the augmented MultiWOZ.
It indicates that the generalization capability to unseen knowledge, locale and domain is another decisive factor towards overall performance improvement.

\begin{figure}[t]
  \centering
  \includegraphics[width=\linewidth]{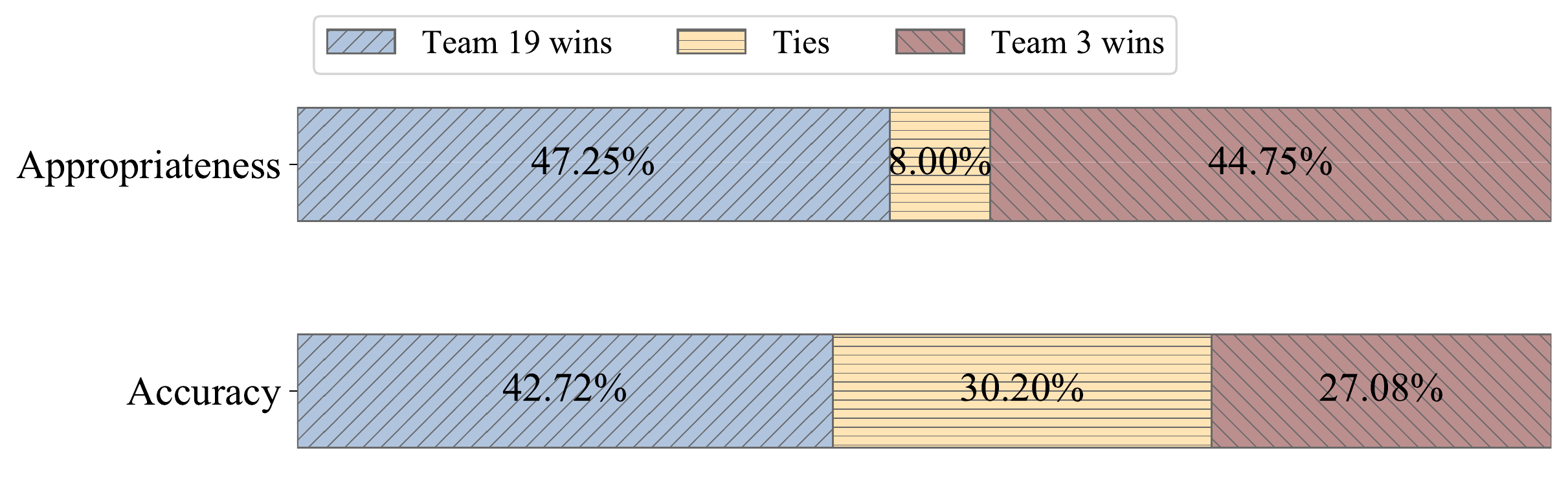}
  \caption{Pairwise human evaluation results between the top-2 system outputs.}
  \label{fig:pair_eval}
\end{figure}

Finally, we performed an additional round of human evaluation to compare between the top-2 systems.
We provided each test instance along with two system outputs; ask three crowd workers to select a more appropriate or accurate response; and took the majority as the final label for the instance.
Figure~\ref{fig:pair_eval} shows that both systems have comparable appropriateness to each other, but Team 19's system generates more accurate responses than Team 3's system.

\section{Conclusions}
We presented the official evaluation results of the Beyond Domain APIs: Task-oriented Conversational Modeling with Unstructured Knowledge Access Track in DSTC9.
This challenge track addressed the new conversational modeling tasks towards frictionless task-oriented dialogues by incorporating unstructured knowledge.
A total of 24 teams participated with an overall number of 105 entries submitted.
From the evaluation results, we learned the following three key factors to achieve high performance in the target tasks:
ensemble of different large-scale pretrained language models,
improved knowledge selection capability,
and better generalization into unseen data.

\bibliography{dstc9_track1}

\begin{thebibliography}{16}
\providecommand{\natexlab}[1]{#1}
\providecommand{\url}[1]{\texttt{#1}}
\providecommand{\urlprefix}{URL }
\expandafter\ifx\csname urlstyle\endcsname\relax
  \providecommand{\doi}[1]{doi:\discretionary{}{}{}#1}\else
  \providecommand{\doi}{doi:\discretionary{}{}{}\begingroup
  \urlstyle{rm}\Url}\fi

\bibitem[{Clark et~al.(2020)Clark, Luong, Le, and Manning}]{clark2020electra}
Clark, K.; Luong, M.-T.; Le, Q.~V.; and Manning, C.~D. 2020.
\newblock Electra: Pre-training text encoders as discriminators rather than
  generators.
\newblock \emph{arXiv preprint arXiv:2003.10555} .

\bibitem[{Dinan et~al.(2018)Dinan, Roller, Shuster, Fan, Auli, and
  Weston}]{dinan2018wizard}
Dinan, E.; Roller, S.; Shuster, K.; Fan, A.; Auli, M.; and Weston, J. 2018.
\newblock Wizard of wikipedia: Knowledge-powered conversational agents.
\newblock \emph{arXiv preprint arXiv:1811.01241} .

\bibitem[{Dong et~al.(2019)Dong, Yang, Wang, Wei, Liu, Wang, Gao, Zhou, and
  Hon}]{dong2019unified}
Dong, L.; Yang, N.; Wang, W.; Wei, F.; Liu, X.; Wang, Y.; Gao, J.; Zhou, M.;
  and Hon, H.-W. 2019.
\newblock Unified language model pre-training for natural language
  understanding and generation.
\newblock In \emph{Advances in Neural Information Processing Systems},
  13063--13075.

\bibitem[{Eric et~al.(2019)Eric, Goel, Paul, Sethi, Agarwal, Gao, and
  Hakkani-Tur}]{eric2019multiwoz}
Eric, M.; Goel, R.; Paul, S.; Sethi, A.; Agarwal, S.; Gao, S.; and Hakkani-Tur,
  D. 2019.
\newblock Multiwoz 2.1: Multi-domain dialogue state corrections and state
  tracking baselines.
\newblock \emph{arXiv preprint arXiv:1907.01669} .

\bibitem[{Galley et~al.(2019)Galley, Brockett, Gao, Gao, and
  Dolan}]{galley2019grounded}
Galley, M.; Brockett, C.; Gao, X.; Gao, J.; and Dolan, B. 2019.
\newblock Grounded response generation task at DSTC7.
\newblock In \emph{Proceedings of the AAAI-19 Workshop on Dialog System
  Technology Challenges}.

\bibitem[{Goel and Byrne(2000)}]{goel2000minimum}
Goel, V.; and Byrne, W.~J. 2000.
\newblock Minimum Bayes-risk automatic speech recognition.
\newblock \emph{Computer Speech \& Language} 14(2): 115--135.

\bibitem[{Gopalakrishnan et~al.(2019)Gopalakrishnan, Hedayatnia, Chen,
  Gottardi, Kwatra, Venkatesh, Gabriel, and
  Hakkani-T{\"u}r}]{gopalakrishnan2019topical}
Gopalakrishnan, K.; Hedayatnia, B.; Chen, Q.; Gottardi, A.; Kwatra, S.;
  Venkatesh, A.; Gabriel, R.; and Hakkani-T{\"u}r, D. 2019.
\newblock Topical-Chat: Towards Knowledge-Grounded Open-Domain Conversations.
\newblock \emph{Proc. Interspeech 2019} 1891--1895.

\bibitem[{Gopalakrishnan et~al.(2020)Gopalakrishnan, Hedayatnia, Wang, Liu, and
  Hakkani-T{\"u}r}]{gopalakrishnan2020neural}
Gopalakrishnan, K.; Hedayatnia, B.; Wang, L.; Liu, Y.; and Hakkani-T{\"u}r, D.
  2020.
\newblock Are Neural Open-Domain Dialog Systems Robust to Speech Recognition
  Errors in the Dialog History? An Empirical Study.
\newblock \emph{Proc. Interspeech 2020} 911--915.

\bibitem[{Kim et~al.(2020)Kim, Eric, Gopalakrishnan, Hedayatnia, Liu, and
  Hakkani-Tur}]{kim-etal-2020-beyond}
Kim, S.; Eric, M.; Gopalakrishnan, K.; Hedayatnia, B.; Liu, Y.; and
  Hakkani-Tur, D. 2020.
\newblock Beyond Domain {API}s: Task-oriented Conversational Modeling with
  Unstructured Knowledge Access.
\newblock In \emph{Proceedings of the 21th Annual Meeting of the Special
  Interest Group on Discourse and Dialogue}, 278--289. 1st virtual meeting:
  Association for Computational Linguistics.
\newblock \urlprefix\url{https://www.aclweb.org/anthology/2020.sigdial-1.35}.

\bibitem[{Liu et~al.(2019)Liu, Ott, Goyal, Du, Joshi, Chen, Levy, Lewis,
  Zettlemoyer, and Stoyanov}]{liu2019roberta}
Liu, Y.; Ott, M.; Goyal, N.; Du, J.; Joshi, M.; Chen, D.; Levy, O.; Lewis, M.;
  Zettlemoyer, L.; and Stoyanov, V. 2019.
\newblock RoBERTa: A Robustly Optimized BERT Pretraining Approach.

\bibitem[{Nogueira and Cho(2019)}]{Nogueira2019PassageRW}
Nogueira, R.; and Cho, K. 2019.
\newblock Passage Re-ranking with BERT.
\newblock \emph{ArXiv} abs/1901.04085.

\bibitem[{Radford et~al.(2019)Radford, Wu, Child, Luan, Amodei, and
  Sutskever}]{radford2019language}
Radford, A.; Wu, J.; Child, R.; Luan, D.; Amodei, D.; and Sutskever, I. 2019.
\newblock Language models are unsupervised multitask learners .

\bibitem[{Spearman(1961)}]{spearman1961proof}
Spearman, C. 1961.
\newblock The proof and measurement of association between two things. .

\bibitem[{Wolf et~al.(2019)Wolf, Debut, Sanh, Chaumond, Delangue, Moi, Cistac,
  Rault, Louf, Funtowicz, and Brew}]{Wolf2019HuggingFacesTS}
Wolf, T.; Debut, L.; Sanh, V.; Chaumond, J.; Delangue, C.; Moi, A.; Cistac, P.;
  Rault, T.; Louf, R.; Funtowicz, M.; and Brew, J. 2019.
\newblock HuggingFace's Transformers: State-of-the-art Natural Language
  Processing.
\newblock \emph{ArXiv} abs/1910.03771.

\bibitem[{Zhang et~al.(2020)Zhang, Sun, Galley, Chen, Brockett, Gao, Gao, Liu,
  and Dolan}]{zhang-etal-2020-dialogpt}
Zhang, Y.; Sun, S.; Galley, M.; Chen, Y.-C.; Brockett, C.; Gao, X.; Gao, J.;
  Liu, J.; and Dolan, B. 2020.
\newblock {DIALOGPT} : Large-Scale Generative Pre-training for Conversational
  Response Generation.
\newblock In \emph{Proceedings of the 58th Annual Meeting of the Association
  for Computational Linguistics: System Demonstrations}, 270--278. Online:
  Association for Computational Linguistics.
\newblock \doi{10.18653/v1/2020.acl-demos.30}.
\newblock \urlprefix\url{https://www.aclweb.org/anthology/2020.acl-demos.30}.

\bibitem[{Zhou et~al.(2018)Zhou, Young, Huang, Zhao, Xu, and
  Zhu}]{zhou2018commonsense}
Zhou, H.; Young, T.; Huang, M.; Zhao, H.; Xu, J.; and Zhu, X. 2018.
\newblock Commonsense Knowledge Aware Conversation Generation with Graph
  Attention.
\newblock In \emph{IJCAI}, 4623--4629.

\end{thebibliography}
\end{document}